%% file: main.tex
\documentclass[letterpaper, 10 pt, conference]{ieeeconf}  

\usepackage{xcolor}

\usepackage{amsmath,amssymb}
\DeclareMathOperator*{\argmax}{arg\,max}

\usepackage{graphics}
\usepackage{graphicx}
\usepackage{multirow}
\usepackage{float}

\IEEEoverridecommandlockouts                              

\makeatletter
\let\NAT@parse\undefined
\makeatother
\usepackage[numbers,sort&compress]{natbib}
\usepackage{hyperref}
\usepackage{graphicx}

\title{\LARGE {\bf
OmniHang:} Learning to Hang Arbitrary Objects using\\ Contact Point Correspondences and  Neural Collision Estimation}

\author{Yifan You$^{1*}$, Lin Shao$^{2*}$,  Toki Migimatsu$^{2}$ and Jeannette Bohg$^{2}$
\thanks{$^*$The authors contributed equally.}
\thanks{$^{1}$Yifan You is with the Department of Computer Science, University of California, Los Angeles, CA, USA.
        {\tt\footnotesize [harry473417@ucla.edu]}}
\thanks{$^{2}$Lin Shao, Toki Migimatsu and Jeannette Bohg are with the Stanford Artificial Intelligence Lab (SAIL), Stanford University, CA, USA. {\tt\footnotesize [lins2,takatoki,bohg]
@stanford.edu}}%
\thanks{Toyota Research Institute (TRI) provided funds to assist the authors with their research but this article solely reflects the opinions and conclusions of its authors and not TRI or any other Toyota entity.}
}

\begin{document}

\maketitle
\thispagestyle{empty}
\pagestyle{empty}

\begin{abstract}
\input{tex/abs.tex}\label{sec:abstract}
\end{abstract}

\section{Introduction}
\input{tex/intro.tex}\label{sec:intro}

\section{Related Work}
\input{tex/relatedwork.tex}\label{sec:relatedwork}

\section{Method}
\input{tex/tech.tex}

\section{Dataset}\label{sec:dataset}
\input{tex/data.tex}

\section{Experiments}\label{sec:exp}
\input{tex/exp.tex}

\section{Conclusion}
\input{tex/con.tex}
\label{sec:conclusion}

{\small
\bibliographystyle{IEEEtranN}
\bibliography{references}
}

\end{document}

%% file: tex/abs.tex
In this paper, we explore whether a robot can learn to hang arbitrary objects onto a diverse set of supporting items such as racks or hooks. Endowing robots with such an ability has applications in many domains such as domestic services, logistics, or manufacturing. Yet, it is a challenging manipulation task due to the large diversity of geometry and topology of everyday objects. In this paper, we propose a system that takes partial point clouds of an object and a supporting item as input and learns to decide where and how to hang the object stably. Our system learns to estimate the contact point correspondences between the object and supporting item to get an estimated stable pose. We then run a deep reinforcement learning algorithm to refine the predicted stable pose. Then, the robot needs to find a collision-free path to move the object from its initial pose to stable hanging pose. To this end, we train a neural network based collision estimator that takes as input partial point clouds of the object and supporting item. We generate a new and challenging, large-scale, synthetic dataset annotated with stable poses of objects hung on various supporting items and their contact point correspondences. In this dataset, we show that our system is able to achieve a 68.3\% success rate of predicting stable object poses and has a 52.1\% F1 score in terms of finding feasible paths. Supplemental material and videos are available on our project webpage~\href{https://sites.google.com/view/hangingobject}{https://sites.google.com/view/hangingobject}.

%% file: tex/intro.tex
Hanging objects is a common daily task. When cleaning a messy bedroom, we may want to hang our hat, bag, or clothes on racks. When arranging a cluttered kitchen, we may want to hang pans or spatulas on hooks to save space. When organizing a workshop, we may want to hang various tools on a pegboard so that they can be easily found. Endowing robots with the ability to autonomously hang a diverse set of objects onto arbitrary supporting items has applications in many domains such domestic services, logistics, or manufacturing. However, the large diversity of geometry and topology in everyday objects makes this a challenging manipulation task. In this paper, we enable a robot to decide where and how to hang arbitrary objects, a task that requires reasoning about contacts and support relationships between two objects.

\begin{figure}[tb!]
 \centering
 \includegraphics[width=\linewidth]{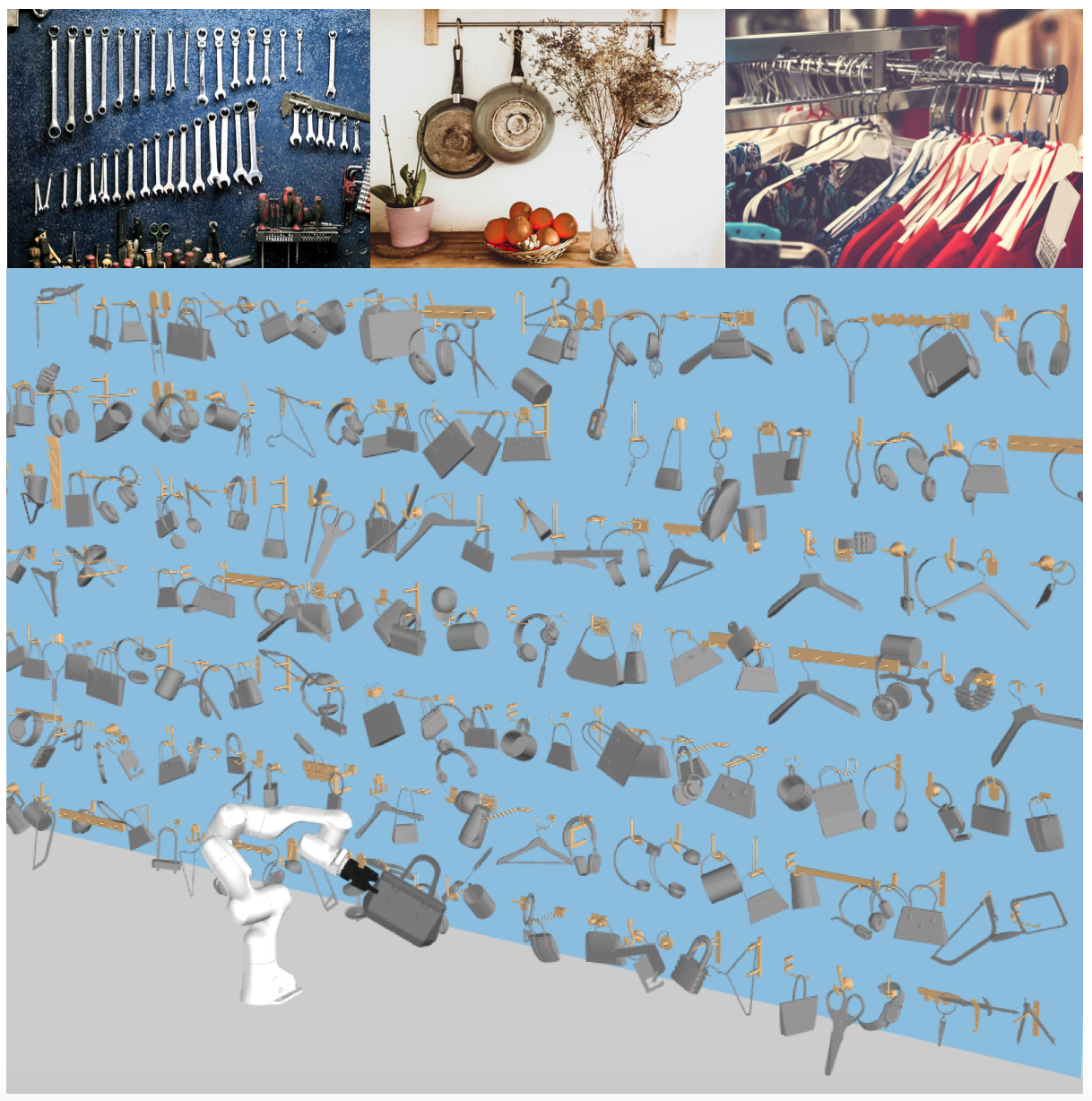}
 \caption{Hanging objects is a common daily task. Our system helps robots learn to hang arbitrary objects onto a diverse set of supporting items such as racks and hooks. All hanging poses rendered here are outputs of our proposed pipeline on object-supporting item pairs unseen during training.}
\label{fig:teaser}
\vspace{-4mm}
\end{figure}

Unlike manipulation tasks such as pick-and-place or peg insertion, hanging objects onto supporting items has not received much attention. \citet{finn2016deep,levine2016end} propose end-to-end learning frameworks for various manipulation tasks, including hanging a rope on a hook and placing a clothes hanger on a rack. These works learn policies that are specific to these pairs of objects and hooks. \citet{jiang2012learning} use Support Vector Machines with hand-designed features to place known objects on dish racks and drawers. While they briefly test hanging objects on hooks, only 40\% of their top 5 hanging proposals are valid. ~\citet{Manuelli2019kPAMKA} demonstrate the ability to hang mugs onto racks by using semantic object keypoints to identify mug-specific geometric features such as the handle or bottom. However, these keypoints are learned from manually annotated data, which can be difficult to scale up to a wide variety of objects. 

In this work, we present a system that takes as input partial point clouds of an object and a supporting item, and addresses the hanging task in two steps:
\begin{enumerate}
\item  \textbf{Where to hang}: Use contact point correspondences to decide where an object should be hung onto its supporting item. We posit that contact point correspondences provide a compact representation of inter-object support and contact relationships. This helps the model to generalize to objects of different categories. 

\item \textbf{How to hang}: Use a neural collision estimator to find a feasible motion plan to hang an object even if it is only partially observed from a depth camera. Prior knowledge of object geometries acquired from data can help the collision estimator to predict collisions from partial observations~\cite{fan2017point,lin2017learning}. 
\end{enumerate}

Our primary contributions are: (1) proposing a contact point matching representation for object manipulation tasks and applying it to learn how to hang arbitrary objects (2) proposing a neural motion planning algorithm to find a collision-free path under partial observation, and (3) generating a large-scale annotated dataset for hanging objects on hooks, racks, or other supporting items. In extensive quantitative experiments, we demonstrate the effectiveness of our proposed method.

%% file: tex/relatedwork.tex
\subsection{Keypoint Representations for Robotic Manipulation}
In robotic manipulation tasks, keypoints are used to provide functional information about the environment and objects. \citet{finn2016deep,levine2016end} use keypoints as intermediate representations of the environment to help learn visuomotor policies that map images to torques to perform manipulation tasks. \citet{qin2020keto} present a framework of learning keypoint representations for tool-based manipulation tasks. The keypoints are divided into tool keypoints and environment keypoints. Tool keypoints are learned from robot interactions. Environment keypoints are predefined to characterize the target position and target force direction in a tool-manipulation task.

The work most related to ours is kPAM~\cite{Manuelli2019kPAMKA}, which represents objects with category-level semantic keypoints. Given the target positions of predefined keypoints, kPAM solves for an optimal transformation to match these predicted correspondences. However, kPAM requires handcrafted 3D semantic keypoints of objects. For example, in order to hang mugs on a mug tree, ~\citet{Manuelli2019kPAMKA} define three mug keypoints to be the top center, bottom center, and handle center. They then design a policy to bring the handle center keypoint to a predefined target point on a mug tree branch. The high cost of manual annotation and the difficulty of handcrafting keypoints and correspondences make this approach difficult to scale up to a wide variety of objects and supporting items. 
 
Our work focuses on learning a general, class-agnostic manipulation model. We apply our pipeline to learn to hang arbitrary objects onto arbitrary supporting items. Our work uses contact points as keypoints, which removes the need to manually define target positions of semantic keypoints, since a contact point on one object must always be aligned with its corresponding contact point on the other object. Using contact points as keypoints also allows us to obtain ground truth annotations of keypoints from simulation by simply querying which points are in contact. Unlike the keypoint detection networks in kPAM, which output a fixed, category-specific number of keypoints, we propose a novel network architecture that proposes many keypoint predictions and ranks them to produce a varying number of keypoints depending on the geometry of two objects.

\subsection{Object Placement}
Pick-and-place is one of the most common tasks in robotic manipulation. Picking objects, or grasping, has attracted great attention in robotics. For a broader review of the field on data-driven grasp synthesis, we refer to~\cite{sahbani2013,bohg2014}. In contrast, object placement, which is the process of deciding where and how to place an object, has received considerably less attention. Most works in object placement are restricted to placing objects on flat horizontal surfaces such as tables~\cite{schuster2010perceiving} or shelves~\cite{edsinger2006manipulation}. ~\citet{jiang2012learning} use Support Vector Machines with hand-designed features to place known objects on dish racks and drawers with a success rate of 98\% and for new objects of 82\%. While the authors briefly test hanging objects on hooks, only 40\% of their top 5 hanging proposals are valid. \citet{finn2016deep,levine2016end} propose end-to-end learning frameworks for various manipulation tasks, including hanging a rope on a hook and placing clothes hangers on a rack. While these works learn the hanging task for specific object-hook pairs, our work learns to hang arbitrary objects onto arbitrary supporting items. 

\subsection{Motion Planning under Partial Observability}
Motion planning finds collision-free paths to bring an object from an initial pose to a goal pose. Typically, motion planning assumes full knowledge of object geometries. In the real world, however, we often do not have access to this information. In our environment, we assume we have depth cameras that only give partial point clouds of objects.

Only a few approaches have extended motion planning to handle partial observability. \citet{agha2014firm,bry2011rapidly} extend sampling-based road map and tree planners to beliefs~(distributions over states). \citet{garrett2020online} perform deterministic cost-sensitive planning in the space of hybrid belief states to select likely-to-succeed observation actions and continuous control actions. While belief state planning can give rise to robust searching or information gathering policies, they are computationally expensive.

To mitigate the computational requirement of motion planning under partial observability, we propose integrating any standard sampling-based motion planner~\cite{kuffner2000rrt,kavraki1996probabilistic} with a neural collision checker that predicts the collision status of objects given their partial point clouds from depth cameras. 

%% file: tex/tech.tex
\begin{figure*}[ht!]
\centering
\includegraphics[width=0.95\linewidth]{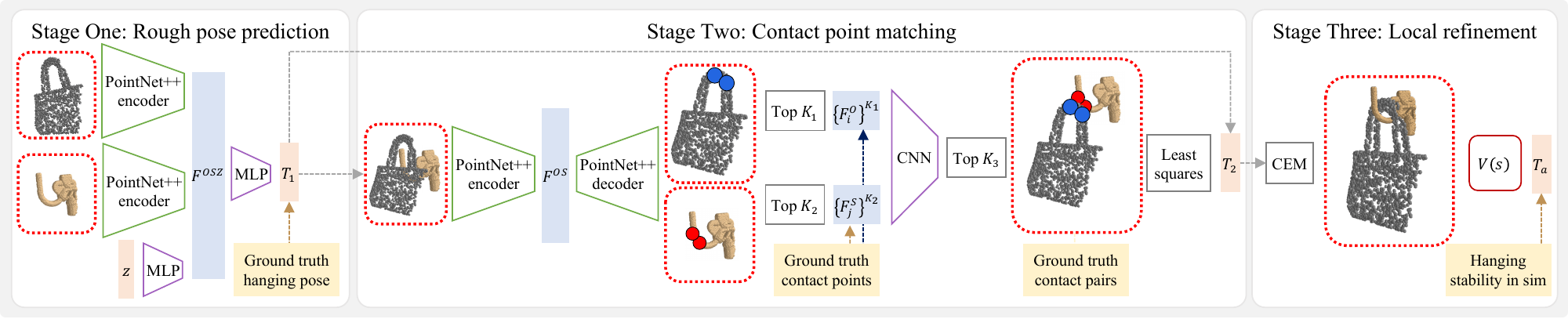}
\caption{Goal pose prediction overview. The model takes in partial point clouds of the object and supporting item, and produces a hanging goal pose for the object over three stages. Each stage is trained individually using ground truth data from simulation. Dashed grey lines represent inputs to functions, and yellow arrows represent supervision signal. Blue and red points are the predicted contact points on object and supporting item, respectively.}\label{fig::pipeline}
\vspace{-3mm}
\end{figure*}

\subsection{Problem Definition}
We consider the problem of a robot hanging a grasped object onto an arbitrary supporting item.  The observations of the object and supporting item are 3D point clouds consisting of M and N points from the robot's RGB-D camera, denoted as $\{P^{O}_i\}^M \in \mathcal{R}^{M\times3}$ and $\{P^S_j\}^N \in \mathcal{R}^{N\times3}$, respectively. We assume that the object and supporting item have already been segmented from the depth camera data.

Given the object's point cloud $\{P^{O}_i\}^M$ and the supporting item's point cloud $\{P^S_j\}^N$, our model first estimates a 6D goal pose for the object. This is the pose in which the object should be stably hung on the supporting item. After our model estimates the goal pose, our method finds a feasible motion plan to move the object from its initial pose to the predicted goal pose.

The following subsections describe the two modules for hanging an object: goal pose prediction (where to hang) and path planning (how to hang).

\subsection{Goal Pose Prediction}
Hanging objects is a complex task that requires a precise goal pose; a slight error may easily result in the object falling down or penetration between the object and its supporting item. Thus, we adopt a three-stage pipeline shown in Fig.~\ref{fig::pipeline} to sequentially estimate and refine the prediction. These stages are explained in detail below. To briefly summarize this pipeline, first, our model takes in partial point clouds of an object and supporting item and outputs a rough initial estimate of a stable hanging pose for the object. Second, based on this initial estimated goal pose, our model predicts contact points between the object and supporting item and aligns them to produce an updated object goal pose. Finally, the updated pose is further refined using reinforcement learning.
We evaluate the importance of each stage through ablation studies in Sec.~\ref{sec:exp}.

\subsubsection{\textbf{Stage One: Rough Pose Prediction}}
Given the point clouds of the object and supporting item, we first use two PointNet++~\cite{qi2017pointnet++} encoders to extract their features denoted as $F^O$ and $F^S$, respectively. There may be a distribution over possible poses to stably hang an object on the supporting item. To encourage our model to explore this distribution, rather than converging to a single solution, we add random noise to our model, inspired by the approach used in Generative Adversarial Networks~\cite{goodfellow2014generative}. This noise is created by first sampling a random variable $z$ from a normal distribution $\mathcal{N}$, and then feeding $z$ into fully connected layers to generate a random noise vector $F^R$. We concatenate $F^0$, $F^S$, and $F^R$ to produce a single feature vector $F^{OSR}$. We then use fully connected layers to decode $F^{OSR}$ to output a 6D pose denoted as ${}^1{\hat{\mathcal{T}}}$ (the top left index stands for Stage One) as a rough initial estimate of a stable hanging pose.

Given the same object point cloud and supporting item point cloud and different sampled $\{z_{l=1}^Z\}$ values, our model outputs $Z$ poses denoted as $\{^1\hat{\mathcal{T}}_{l=1}^Z\}$. For a pair of the object and supporting item, our dataset described in Sec.~\ref{sec:dataset} contains a ground truth list of stable 6D poses denoted as $\{^1\mathcal{T}_{k=1}^D\}$. We defined a loss $\mathcal{L}_{M}$ to train our model such that the set of predicted poses $\{^1\hat{\mathcal{T}}_{l=1}^Z\}$ are close to the set of ground truth 6D poses $\{^1\mathcal{T}_{k=1}^D\}$. Note that $\{^1\hat{\mathcal{T}}_{l=1}^Z\}$ are the outputs based on the input minibatch. Entries in the minibatch only vary in $z_l$. Therefore we are forcing the model to implicitly learn the distribution by minimizing the loss $\mathcal{L}_M$. We define a loss denoted as $L$ to measure the difference between one predicted pose and one ground truth pose. The 6D pose is composed of 3D translation and 3D axis-angle vector. For orientation, we adopt the axis angle representation, which has been shown to be effective for pose prediction task~\cite{8411477}. $L$ is a linear combination of L2 losses for 3D position vector and 3D axis angle vector.
\begin{align}
\begin{split}
&\mathcal{L}_M(\{^1\hat{\mathcal{T}}_{l=1}^Z\}, \{^1\mathcal{T}_{k=1}^D\}) = \\
&\qquad\qquad \sum_{l=1}^Z \min_k L(^1\hat{\mathcal{T}}_l, \,^1\mathcal{T}_k) + \sum_{k=1}^D \min_l L(^1\hat{\mathcal{T}}_l, \,^1\mathcal{T}_k)
\end{split}
\end{align}

\subsubsection{\textbf{Stage Two: Contact Point Matching}}
For each predicted pose $\hat{\mathcal{T}}_l$ from Stage One, our model transforms the object point cloud to $\{^1\hat{\mathcal{T}}_lP_i^O\}^M$ based on the predicted pose. This stage predicts which points on the transformed object and supporting item should be in contact for the stable hanging pose and updates the goal pose accordingly. First, we augment these two point clouds $\{^1\hat{\mathcal{T}}_lP_i^O\}^M$ and $\{P_j^S\}^N$ by adding extra 1D arrays of $\{1\}^M$ and $\{0\}^N$ along the XYZ dimension, respectively. Our model combines these two point clouds into one point cloud which has a shape of $(M+N,4)$. The combined point cloud is fed into a PointNet++ encoder to produce a feature vector $F^{OS}$. The feature vector is then fed into a PointNet++ decoder to output a score $\hat{s}_i$ for each point on the object and supporting item indicating whether this point should be in contact. Our model selects the top $K_1$ and $K_2$ points of the object and supporting item, respectively.

Next, the model predicts the contact point correspondences $\{C(i,j)\}$ between the selected contact point sets $\{p^O_i\}^{K_1}$ and $\{p^S_j\}^{K_2}$. Each feature vector $\{F^{OS}_i\}^{K_1}$ and $\{F^{OS}_j\}^{K_2}$ associated with the selected points is a vector of size $W$. Our model performs a pairwise copy and concatenation to produce a combined feature map of dimension $(K_1, K_2, W+W)$. The feature map is then sent to 1D convolution layers and softmax layer to get the final score matrix of dimension $(K_1, K_2)$. Each element $C(u,v)$ of the matrix represents the probability $\hat{y}_{(u,v)}$ that $p^O_u$ and $p^S_v$ are a valid contact point pair when the object is hung stably on the supporting item. Then our model ranks these probabilities and selects the top $K_3$ with probability larger than a given threshold $\delta$.

Finally, given the correspondences between the object and supporting item, our model solves for the optimal translation $^2\hat{T}$ to minimize the total distance between the paired contact points of the object and supporting item, using the orientation $^1\hat{R}$ predicted by Stage One. The updated goal pose from Stage Two is then $^2\hat{\mathcal{T}} = \begin{bmatrix}^1\hat{R} & ^2\hat{T}\end{bmatrix}$. We do not optimize the orientation at this stage because the contact points may not restrict the degrees of freedom enough to determine the orientation. For example, if an object makes contact with its supporting item at only one point, then it is free to rotate about that point.

We divide the training of this stage into two parts. First, we train the PointNet++ decoder, which predicts $\hat{s}_i$ for each point on the object and supporting item indicating whether the point should be in contact. We can obtain ground truth contact point labels $s_i$ from the dataset of simulated stable hanging poses as described in Sec.~\ref{sec:dataset}. We formulate the prediction to be a point-wise regression problem and adopt the L2 loss $\mathcal{L}_2=\|\hat{s}_i - s_i\|$.

Second, we train the contact point correspondence network, which takes the $K_1$ and $K_2$ selected contact points on the object and supporting item, respectively, and outputs a $K_1 \times K_2$ matrix indicating the probability of each pair of points being in contact with each other. We formulate contact point correspondence as a binary classification problem and adopt the cross entropy loss $\mathcal{L}_c$. Since we are selecting the top $K_3$ point pairs from the probability matrix $\hat{y}_{(u,v)}$, we add the ranking loss $\mathcal{L}_r$ to encourage points with positive labels to have a higher ranking than points with negative labels. This ranking loss is a variant of ListNet~\cite{cao2007learning,shao2020unigrasp} and defined as follows:
\begin{equation}
    \mathcal{L}_r = -\sum_{(u,v)}^{K_1\times K_2} y_{(u,v)} \log(\frac{\exp(\hat{y}_{(u,v)})}{\sum_{(u,v)}^{K1 \times K2} \exp(\hat{y}_{(u,v)})})
\end{equation}
The total loss of assigning contact point correspondences is $\mathcal{L}_2 = \mathcal{L}_c + \mathcal{L}_r$. 

\subsubsection{\textbf{Stage Three: Local Refinement}}\label{sec:localR}
The last stage is to perform a final refinement of the predicted goal pose. We formulate the refinement process as a reinforcement learning~(RL) problem and learn a policy to compensate for the errors produced by the first two stages, assuming that the first two stages can produce a reasonable estimate for a stable goal pose. The RL algorithm learns a value function $\mathcal{V}$ that models the expected feasibility of nearly-correct goal poses. At test time, we probe this value function with CEM~\cite{CEM} to find a locally optimal goal pose.

We first apply the transformation $^2\hat{\mathcal{T}}$ predicted by Stage Two to the object point cloud $\{P^{O}_i\}^M$ to get a point cloud $\{\tilde{P}^{O}_i\}^M$. The initial state $s_0$ for RL training is $\{\tilde{P}^{O}_i\}^M$ with the supporting item point cloud $\{P^S_j\}^M$. Through this initialization, we limit the amount of exploration required by the RL algorithm, making learning easier.

The action $a$ is a 6D transformation $\mathcal{T}_a$ which transforms the object point cloud $\{\tilde{P}^{O}_i\}^M$ into $\{\mathcal{T}_a\tilde{P}^{O}_i\}^M$. The new state $s$ is the transformed object point cloud along with the supporting item point cloud $(\{\mathcal{T}_a\tilde{P}^{O}_i\}^M,\{P^{S}_j\}^M)$. If the transformed object hangs stably on the supporting item in simulation, the reward $r$ is one. Otherwise, $r$ is zero. 

To solve this RL problem, we train a value function $\mathcal{V}(s)$ based on PointNet++ to approximate the expected reward of each state $s = (\{\mathcal{T}_a\tilde{P}^{O}_i\}^M,\{P^{S}_j\})$. We collect the transition data $\{(a,r,s)\}$ and train the value model $\mathcal{V}(s)$ by minimizing the following loss:
\begin{equation}
    \mathcal{L}_v = \|\mathcal{V}(\{\mathcal{T}_a\tilde{P}^{O}_i\}^M,\{P^{S}_j\}^M) - r\|
\vspace{-1mm}
\end{equation}
When selecting the action $a$, we run a derivative-free optimization method CEM~\cite{CEM} to search within the 6D pose space to find a 6D transformation $\mathcal{T}_a$ associated with the highest score in the value model $\mathcal{V}(s)$ .
\begin{equation}
    a^* = \argmax_a{\mathcal{V}(\{\mathcal{T}_a\tilde{P}^{O}_i\}^M,\{P^{S}_j\}^M})
\end{equation}


\begin{figure}
\centering
\includegraphics[width=0.92\columnwidth]{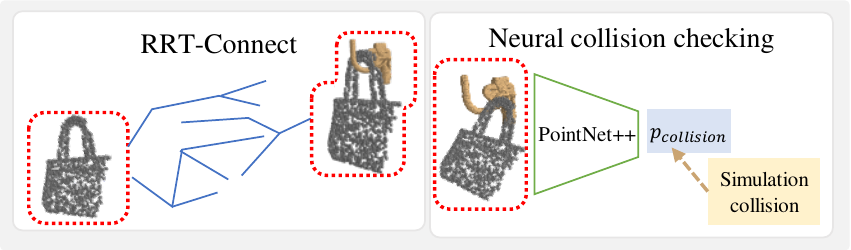}
\caption{Our neural collision checker acts as a state validity checker, and thus integrates with any sampling-based motion planner. We use RRT-Connect to find a collision-free path to hang the object. Neural collision checking takes in partial point clouds of the object and supporting item and outputs a score indicating whether the two objects are colliding.}\label{fig:pathplanning}
\vspace{-4mm}
\end{figure}

\vspace{-4mm}
\subsection{Neural Path Planning}
Given a predicted goal pose for the object, the next step is to find a collision-free path to move the object from its initial pose to the predicted hanging pose. Our approach is visualized in Figure~\ref{fig:pathplanning}. We use RRT-Connect~\cite{kuffner2000rrt} to find a feasible path, but any sampling-based planner can be used at this step. Sampling-based motion planners ~\cite{kuffner2000rrt,kavraki1996probabilistic} require collision estimation which typically requires full information about the geometry of objects and environment. However, we start from the realistic scenario in which we perceive the objects and supporting item through a depth camera and therefore only observe partial point clouds. We propose to train a deep neural network to estimate the collision status of two point clouds in a specific relative pose.

The neural collision estimator takes as inputs the two points clouds and outputs a score representing collision probability of the two point clouds. We formulate collision estimation as a binary classification problems. We automatically gather ground truth positive and negative training examples from simulation. We apply a standard cross entropy loss for training the model.

%% file: tex/data.tex
\subsection{Generating the Hanging Pose Dataset}
The dataset contains 340 objects and 72 supporting items. For each pair of object/supporting item, the initial hanging poses are generated by sampling different object poses w.r.t the supporting item in PyBullet, and running a forward simulation to see if it falls. In some of the poses, the object cannot be taken off the supporting item. To check this, we apply a constant force to the object for a certain number of timesteps in PyBullet, to see if the object can be taken off. Examples where the object cannot be taken off are discarded. In total, our dataset for pose prediction contains 19,299 pairs of object/supporting item. We split the dataset into 16,195 pairs for training and 3,104 pairs for testing.
\subsection{Auto-Annotating Contact Points}
\vspace{-1mm}
Given a hanging pose, Stage Two of our pipeline requires contact point information for each pose as an additional supervision signal. In simulation, we obtain contact points on both the object and supporting item meshes. For each contact point on the object, we select a neighborhood of points on the object’s point cloud closest to the contact point. We store the points selected on the object and supporting item point clouds together with the contact point correspondences between the object and supporting item.

%% file: tex/exp.tex
Our experiments focus on evaluating the following questions: (1) How crucial is each of the three stages in the pipeline for finding stable hanging poses? (2) How well can neural collision checking with partial observability perform?

\begin{figure}[t]
\centering
\includegraphics[width=0.9\linewidth]{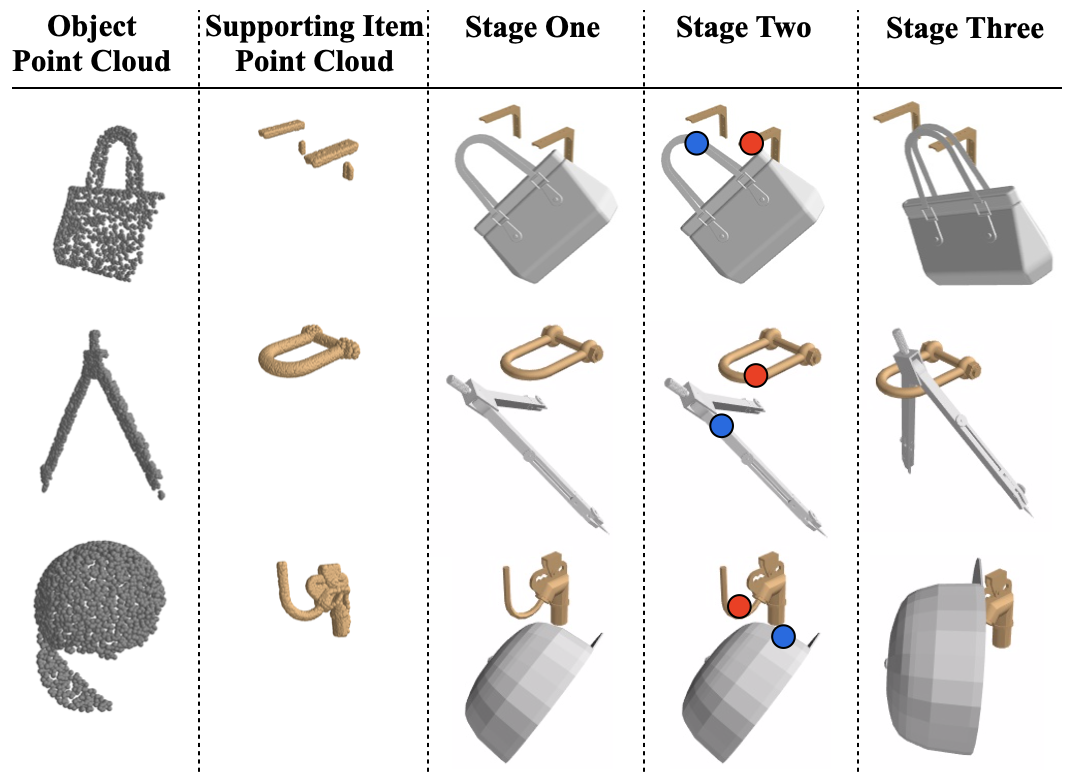}
\caption{Visualizations of intermediate results from all three stages. Our model first outputs a rough pose prediction (Stage One), then predicts contact point correspondences (Stage Two), and finally performs local refinement (Stage Three). Blue and red points are the predicted corresponding contact points on object and supporting item, respectively. After registering the contact correspondences in Stage Two, the red and blue points would be aligned. Only one pair of contact points is shown for visual clarity. Partial point clouds of the object and supporting item are generated from a single camera view in simulation. Meshes are shown for Stages 1--3 instead of partial point clouds for visual clarity.}\label{fig:allstage}
\vspace{-5mm}
\end{figure}

\subsection{Experimental Setup}
Given a pair of object and supporting item, our model predicts a stable hanging pose. For evaluation, we use PyBullet~\cite{coumans2019} to test whether the predicted pose is stable. We load the object at the predicted pose and check if it is collision-free and whether the object falls down to the ground. The resulting stable goal pose is then input to the neural motion planning pipeline that searches for a feasible path. After the planner returns a predicted feasible path, we test whether the path is collision-free in PyBullet.

\subsection{Goal Pose Prediction}
The goal pose prediction pipeline is comprised of three stages that successively update the goal pose: rough goal prediction, contact point matching, and local refinement. To assess the importance of each stage, we evaluate the hanging success rate per ablated baseline and compare to the full pipeline. We also evaluate whether we can learn to hang objects using only RL. For this, we compare the learning speed of Stage Three with and without being provided with a goal pose by the preceding stages. Fig.~\ref{fig:allstage} shows some examples of predicted poses at each stage of our whole pipeline. Note that our dataset contains a rich set of objects and supporting items with diverse geometry and topology as visualized in Fig.~\ref{fig:teaser}.

\subsubsection{\textbf{Hanging success}}
\begin{figure*}[h!]
\small
\begin{tabular}{@{}p{0.13\linewidth}|
 p{0.05\linewidth} p{0.04\linewidth} p{0.04\linewidth} p{0.04\linewidth} p{0.04\linewidth} p{0.06\linewidth} p{0.05\linewidth} p{0.05\linewidth} p{0.05\linewidth} p{0.05\linewidth} p{0.06\linewidth} p{0.05\linewidth}}
\hline \hline 

Methods & Mean & Bag & Cap
& Hanger & Utensil & Headphone
& Knife & Mug & Racquet & Scissors
& Wrench  & Others   \\ \hline
Stage 1 only      & 36.0 & 50.0  & 58.5 & 44.0 & 17.7 & \hspace{1mm} 41.2 & 26.0 & 30.3 & 28.7 & 22.5 & 21.5 & 30.7 \\
Stages 1+2        & 34.2 & 47.4 & 55.7 & 56.0 & 20.1 & \hspace{1mm} 30.2 & 30.8 & 20.7 & 27.6 & 22.1 & 27.9 & 30.0 \\
Stages 1+3        & 56.2 & 63.6 & \textbf{60.8} & 69.1 & 37.8 & \hspace{1mm} 60.9 & 46.2 & 44.4 & 44.8 & 55.4 & 41.9 & 58.1 \\
All Stages (1--3) & \textbf{68.3} & \textbf{78.3} & 60.2 & \textbf{72.2} & \textbf{45.1} & \hspace{1mm} \textbf{80.5} & \textbf{49.0} & \textbf{61.9} & \textbf{49.4} & \textbf{66.2} & \textbf{55.2} & \textbf{71.5}\\
\hline 
\hline
\end{tabular}
\caption{Hanging pose prediction accuracy. We report the mean accuracy for each object category and across all categories. Using all three stages achieves the highest accuracy for all but one object category.}
\label{tab:suctask}
\vspace{-4mm}
\end{figure*}

The first baseline is Stage One alone, which can be considered to be a vanilla pose estimation method. This simply feeds point clouds of the object and supporting item to PointNet++~\cite{qi2017pointnet++} to directly output a 6D hanging pose. The second baseline combines Stages One and Two and skips the refinement Stage Three, and the third baseline combines Stages One and Three and skips the contact point matching stage. We evaluate these baselines by testing the stability of their predicted poses. The results are shown in Fig.~\ref{tab:suctask}. 

While our full pipeline achieves a mean success rate of 68.3\%, Stage One alone achieves only 36.0\%. This underlines the difficulty of hanging an arbitrary object on an arbitrary supporting item. It requires a highly precise prediction of the object's stable pose, and the wide diversity of objects makes this problem challenging. Stages One and Two together achieve 34.2\% which is lower than the performance of Stage One~(36.0\%) alone. The reason for this is that in Stage Two, the object is aligned with the supporting item by minimizing the distance between matched contact points. This alignment might lead to a collision and requires Stage Three to refine the object pose to become feasible. However, simple neglecting Stage Two and only using Stage One and Three leads to a success rate of 56.2\% which is lower than 68.3\% when using all stages. For all but the Cap class, Stage Two improves the performance of the pipeline.

\subsubsection{\textbf{Refinement learning}}
Stage Three is the bottleneck of our entire pipeline in terms of training time. To quantify the importance of the first two stages, we evaluate the learning speed of the RL-based refinement stage with and without initialization from the first two stages. Our hypothesis is that initializing Stage Three with a good estimate of the goal pose will significantly reduce training time. As a baseline, we use Stage Three by itself which equates to a pure deep RL algorithm that learns to output a goal pose given the initial point clouds.

\begin{figure}[htb!]
\centering
\includegraphics[width=0.7\linewidth]{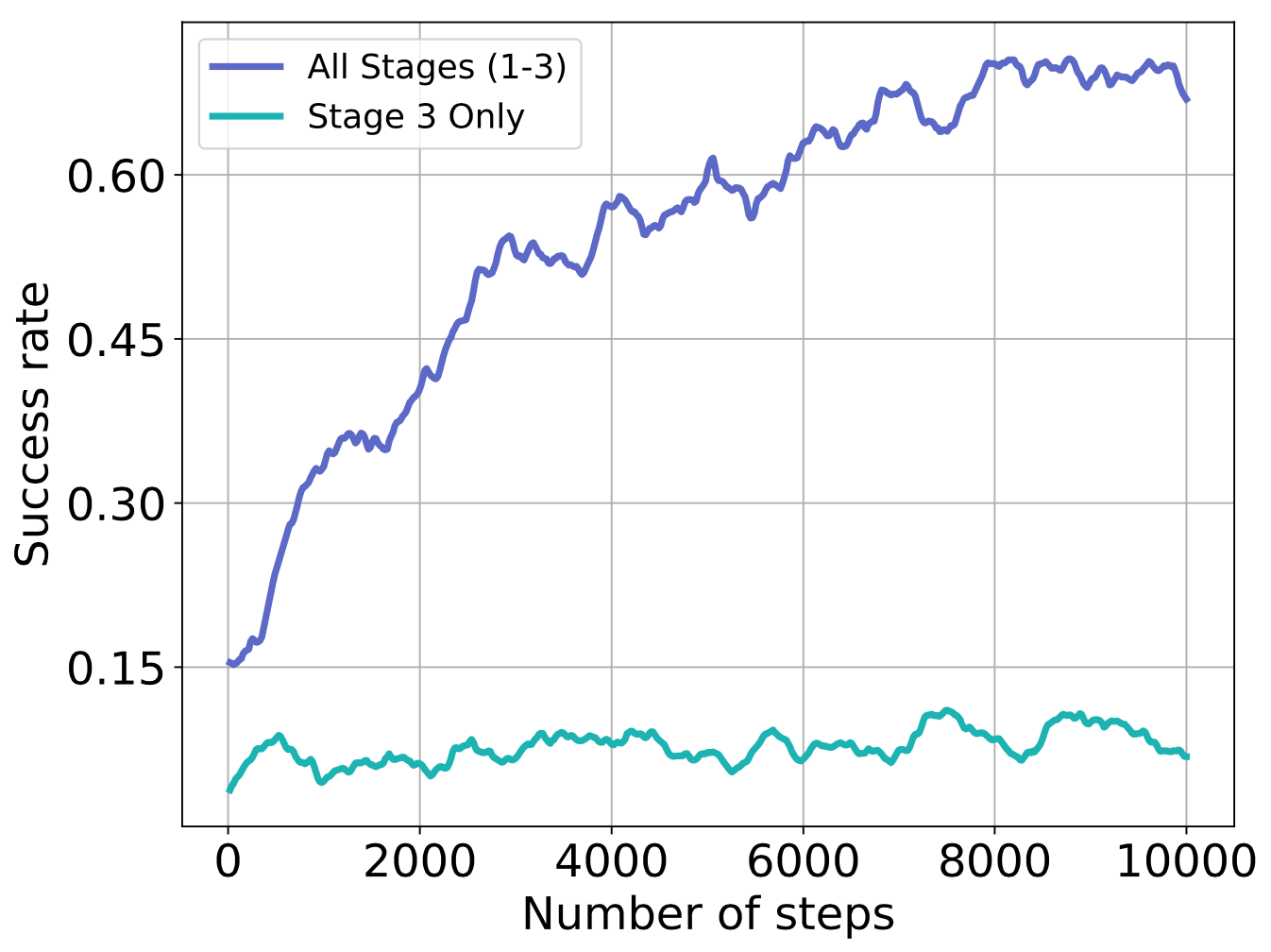}
\vspace{-2mm}
\caption{Learning curves of running RL on 100 randomly sampled pairs of object/supporting item. Stage Three alone uses a uniformly sampled pose to initialize the RL algorithm, while the full pipeline initializes RL with the output of Stage Two. The full pipeline learns faster by a factor of six.}\label{fig::learningCurve}
\end{figure}

We compare RL learning speed on a subset of the training set containing 100 randomly sampled pairs of objects and supporting items. The training curves are shown in Fig.~\ref{fig::learningCurve}. In 10,000 training steps, Stage Three alone achieves a success rate of only 10\%, while initializing with the output of Stage Two achieves 71\%. This result indicates the importance of initializing the RL algorithm with a reasonable solution, in order to minimize the amount of exploration required. Our full pipeline improves the learning speed by a factor of six.

\subsection{Neural Collision Estimation}~\label{sec:hce}
Our neural collision estimator identifies the collision status between objects from their partially observed point clouds. Given a stable pose of the object hung on a supporting item, we place the object at a pre-defined pose far away from the the supporting item, and run each motion planner to move the object to the stable hanging pose on the supporting item. In simulation, we utilize mesh models of objects and supporting items to find these feasible paths. We denote it as ground truth. We compare our neural collision estimator~(\textbf{NCE}) with a heuristic collision estimator~(\textbf{HCE}).

\textbf{Heuristic Collision Estimation~(HCE).}
Given a single point $P^O_i$ on the object's point cloud, denote $P^S_k$ as the nearest point to $P^O_i$ on the supporting item's point cloud. Let $Pn^S_k$ be the outward-pointing normal of $P^S_k$. We observe that when the object is penetrating the supporting item and $P^O_i$ is inside the supporting item's point cloud, $P^O_i$ is typically on the inside halfspace of $P^S_k$ such that $(P^O_i-P^S_k)^T Pn^S_k \leq 0$. Thus, we can use the cosine distance $\frac{(P^O_i-P^S_k)^T}{|P^O_i-P^S_k|} Pn^S_k$ as a measure of how far $P^O_i$ is penetrating the supporting item. We calculate this distance for every point on the object's point cloud, and take the negative of the average as the collision score. If this collision score is above a pre-defined threshold, HCE returns a collision. 

To estimate the outward-facing normals, we train a neural network based on PointNet++~\cite{qi2017pointnet++} which takes partial point clouds as input and outputs an outward-pointing normal vector $\hat{Pn}_i$ for each point. These normals provide features for identifying the interior region of the object.  Given the ground truth point normals $\{Pn_i\}_{i=1}$ gathered from simulation, the training loss of predicted point normals is defined to be
\begin{equation}
    L_{Pn} = -\sum_{i=1}(\hat{Pn}_i^T Pn_i)
\end{equation}

We evaluate each of the motion planning methods using precision and recall. Precision measures the proportion of predicted paths returned by the planner that are collision-free. Recall measures the proportion of all pairs of object and supporting item for which the planner finds a collision-free path. \textbf{HCE} has a precision of 48.2\% and a recall of 32.4\% with a 38.7\% F1 score. \textbf{NCE} achieves a precision of 65.8\% and a recall of 43.2\% with a 52.1\% F1 score. 

\textbf{NCE} outperforms \textbf{HCE} on both precision and recall. It reflects that \textbf{NCE} is able to identify collision based on partial observations but also finds more feasible paths. We believe that motion planning under partial observability remains an open and challenging problem, and data-driven methods have shown promising results. Qualitative results are available on our ~\href{https://sites.google.com/view/hangingobject}{project webpage}.

%% file: tex/con.tex
We present a system that can hang arbitrary objects onto a diverse set of supporting items such as racks and hooks. Our system learns to decide where and how to hang the object stably based on partial point clouds of the object and the supporting item. It predicts stable poses by first predicting contact point correspondences between the object and supporting item to represent their contact relationship. Then our system uses a reinforcement learning algorithm to refine the predicted stable pose. Once the hanging goal pose is established, we use neural network-based collision estimation to find a feasible path to hang the object under partial observability. We demonstrate the effectiveness of our system on a new and challenging, large-scale, synthetic dataset and show that our system is able to achieve a 68.3\% success rate of predicting stable object poses and has a 52.1\% F1 score in terms of finding feasible paths. While we show promising results in simulated environments, we look forward to run our approach on real robot hardware post-COVID. For future work, we would like to apply our method of learning contact point correspondences to a wider range of robotic manipulation tasks such as object assembly and soft object manipulation.